\documentclass[runningheads]{llncs}
\usepackage{graphicx}
\usepackage{amsmath,amssymb} 
\usepackage{color}

\usepackage{graphicx}
\usepackage{url}
\usepackage{caption}
\usepackage{booktabs}
\usepackage{multirow}
\usepackage{ctable}
\usepackage[width=122mm,left=12mm,paperwidth=146mm,height=193mm,top=12mm,paperheight=217mm]{geometry}
\newcommand{\overbar}[1]{\mkern 1.5mu\overline{\mkern-1.5mu#1\mkern-1.5mu}\mkern 1.5mu}

\def\etal{\emph{et al. }}

\begin{document}
\title{MVSNet: Depth Inference for \\Unstructured Multi-view Stereo} 

\titlerunning{MVSNet}
%

\author{
Yao Yao$^{1}$ \and Zixin Luo$^{1}$ \and Shiwei Li$^{1}$ \and Tian Fang$^{2}$ \and  Long Quan$^{1}$
}

%
\authorrunning{Y. Yao, Z. Luo, S. Li, T. Fang, L. Quan}
%

\institute{
$^1$~The Hong Kong University of Science and Technology, \\
\email{\{yyaoag, zluoag, slibc, quan\}@cse.ust.hk} \\
$^2$~Shenzhen Zhuke Innovation Technology (Altizure), \\
\email{fangtian@altizure.com}
}

\maketitle

\begin{abstract}
We present an end-to-end deep learning architecture for depth map inference from multi-view images. In the network, we first extract deep visual image features, and then build the 3D cost volume upon the reference camera frustum via the differentiable homography warping. Next, we apply 3D convolutions to regularize and regress the initial depth map, which is then refined with the reference image to generate the final output. Our framework flexibly adapts arbitrary N-view inputs using a variance-based cost metric that maps multiple features into one cost feature. The proposed MVSNet is demonstrated on the large-scale indoor \textit{DTU} dataset. With simple post-processing, our method not only significantly outperforms previous state-of-the-arts, but also is several times faster in runtime. We also evaluate MVSNet on the complex outdoor \textit{Tanks and Temples} dataset, where our method ranks first before April 18, 2018 without any fine-tuning, showing the strong generalization ability of MVSNet.
\end{abstract}

\section{Introduction}

Multi-view stereo (MVS) estimates the dense representation from overlapping images, which is a core problem of computer vision extensively studied for decades. Traditional methods use hand-crafted similarity metrics and engineered regularizations (e.g., normalized cross correlation and semi-global matching \cite{hirschmuller2008stereo}) to compute dense correspondences and recover 3D points. While these methods have shown great results under ideal Lambertian scenarios, they suffer from some common limitations. For example, low-textured, specular and reflective regions of the scene make dense matching intractable and thus lead to incomplete reconstructions. It is reported in recent MVS benchmarks \cite{aanaes2016large,knapitsch2017tanks} that, although current state-of-the-art algorithms \cite{furukawa2010accurate,vu2012high,galliani2015massively,schonberger2016pixelwise} perform very well on the \emph{accuracy}, the reconstruction \emph{completeness} still has large room for improvement. 

Recent success on convolutional neural networks (CNNs) research has also triggered the interest to improve the stereo reconstruction. Conceptually, the learning-based method can introduce global semantic information such as specular and reflective priors for more robust matching. There are some attempts on the two-view stereo matching, by replacing either hand-crafted similarity metrics \cite{zbontar2016stereo,han2015matchnet,luo2016efficient,hartmann2017learned} or engineered regularizations \cite{seki2017sgm,knobelreiter2017end,kendall2017end} with the learned ones. They have shown promising results and gradually surpassed traditional methods in stereo benchmarks \cite{geiger2012we,menze2015object}. In fact, the stereo matching task is perfectly suitable for applying CNN-based methods, as image pairs are rectified in advance and thus the problem becomes the horizontal pixel-wise disparity estimation without bothering with camera parameters.

However, directly extending the learned two-view stereo to multi-view scenarios is non-trivial. Although one can simply pre-rectify all selected image pairs for stereo matching, and then merge all pairwise reconstructions to a global point cloud, this approach fails to fully utilize the multi-view information and leads to less accurate result. Unlike stereo matching, input images to MVS could be of arbitrary camera geometries, which poses a tricky issue to the usage of learning methods. Only few works acknowledge this problem and try to apply CNN to the MVS reconstruction: SurfaceNet \cite{ji2017surfacenet} constructs the Colored Voxel Cubes (CVC) in advance, which combines all image pixel color and camera information to a single volume as the input of the network. In contrast, the Learned Stereo Machine (LSM) \cite{kar2017learning} directly leverages the differentiable projection/unprojection to enable the end-to-end training/inference. However, both the two methods exploit the volumetric representation of regular grids. As restricted by the huge memory consumption of 3D volumes, their networks can hardly be scaled up: LSM only handles synthetic objects in low volume resolution, and SurfaceNet applies a heuristic divide-and-conquer strategy and takes a long time for large-scale reconstructions. For the moment, the leading boards of modern MVS benchmarks are still occupied by traditional methods \cite{furukawa2010accurate,galliani2015massively,schonberger2016pixelwise}.

To this end, we propose an end-to-end deep learning architecture for depth map inference, which computes one depth map at each time, rather than the whole 3D scene at once. Similar to other depth map based MVS methods \cite{tola2012efficient,campbell2008using,galliani2015massively,schonberger2016pixelwise}, the proposed network, MVSNet, takes one reference image and several source images as input, and infers the depth map for the reference image. The key insight here is the differentiable homography warping operation, which implicitly encodes camera geometries in the network to build the 3D cost volumes from 2D image features and enables the end-to-end training. To adapt arbitrary number of source images in the input, we propose a variance-based metric that maps multiple features into one cost feature in the volume. This cost volume then undergoes multi-scale 3D convolutions and regress an initial depth map. Finally, the depth map is refined with the reference image to improve the accuracy of boundary areas. There are two major differences between our method and previous learned approaches \cite{kar2017learning,ji2017surfacenet}. First, for the purpose of depth map inference, our 3D cost volume is built upon the camera frustum instead of the regular Euclidean space. Second, our method decouples the MVS reconstruction to smaller problems of per-view depth map estimation, which makes large-scale reconstruction possible.

We train and evaluate the proposed MVSNet on the large-scale \textit{DTU} dataset \cite{aanaes2016large}. Extensive experiments show that with simple post-processing, MVSNet outperforms all competing methods in terms of completeness and overall quality. Besides, we demonstrate the generalization power of the network on the outdoor \textit{Tanks and Temples} benchmark \cite{knapitsch2017tanks}, where MVSNet ranks first (before April. 18, 2018) over all submissions including the open-source MVS methods (e.g., COLMAP \cite{schonberger2016pixelwise} and OpenMVS \cite{openMVS}) and commercial software (Pix4D \cite{Pix4D}) without any fine-tuning. It is also noteworthy that the runtime of MVSNet is several times or even several orders of magnitude faster than previous state-of-the-arts.

\section{Related work}

\textbf{MVS Reconstruction.}
According to output representations, MVS methods can be categorized into 1) direct point cloud reconstructions \cite{lhuillier2005quasi,furukawa2010accurate}, 2) volumetric reconstructions \cite{kutulakos2000theory,seitz1999photorealistic,ji2017surfacenet,kar2017learning} and 3) depth map reconstructions \cite{tola2012efficient,campbell2008using,galliani2015massively,schonberger2016pixelwise,yao2017relative}. 
Point cloud based methods operate directly on 3D points, usually relying on the propagation strategy to gradually densify the reconstruction \cite{lhuillier2005quasi,furukawa2010accurate}. As the propagation of point clouds is proceeded sequentially, these methods are difficult to be fully parallelized and usually take a long time in processing. 
Volumetric based methods divide the 3D space into regular grids and then estimate if each voxel is adhere to the surface. The downsides for this representation are the space discretization error and the high memory consumption. 
In contrast, depth map is the most flexible representation among all. It decouples the complex MVS problem into relatively small problems of per-view depth map estimation, which focuses on only one reference and a few source images at a time. Also, depth maps can be easily fused to the point cloud \cite{merrell2007real} or the volumetric reconstructions \cite{newcombe2011kinectfusion}. According to the recent MVS benchmarks \cite{aanaes2016large,knapitsch2017tanks}, current best MVS algorithms \cite{galliani2015massively,schonberger2016pixelwise} are both depth map based approaches. 

\noindent \textbf{Learned Stereo. } 
Rather than using traditional handcrafted image features and matching metrics \cite{hirschmuller2007evaluation}, recent studies on stereo apply the deep learning technique for better pair-wise patch matching. Han \etal \cite{han2015matchnet} first propose a deep network to match two image patches. Zbontar \etal \cite{zbontar2016stereo} and Luo \etal \cite{luo2016efficient} use the learned features for stereo matching and semi-global matching (SGM) \cite{hirschmuller2008stereo} for post-processing. Beyond the pair-wise matching cost, the learning technique is also applied in cost regularization. SGMNet \cite{seki2017sgm} learns to adjust the parameters used in SGM, while CNN-CRF \cite{knobelreiter2017end} integrates the conditional random field optimization in the network for the end-to-end stereo learning. The recent state-of-the-art method is GCNet \cite{kendall2017end}, which applies 3D CNN to regularize the cost volume and regress the disparity by the soft argmin operation. It has been reported in KITTI banchmark \cite{menze2015object} that, learning-based stereos, especially those end-to-end learning algorithms \cite{mayer2016large,knobelreiter2017end,kendall2017end}, significantly outperform the traditional stereo approaches.

\noindent \textbf{Learned MVS. } There are fewer attempts on learned MVS approaches. Hartmann \etal propose the learned multi-patch similarity \cite{hartmann2017learned} to replace the traditional cost metric for MVS reconstruction. 
The first learning based pipeline for MVS problem is SurfaceNet \cite{ji2017surfacenet}, which pre-computes the cost volume with sophisticated voxel-wise view selection, and uses 3D CNN to regularize and infer the surface voxels. 
The most related approach to ours is the LSM \cite{kar2017learning}, where camera parameters are encoded in the network as the projection operation to form the cost volume, and 3D CNN is used to classify if a voxel belongs to the surface. 
However, due to the common drawback of the volumetric representation, networks of SurfaceNet and LSM are restricted to only small-scale reconstructions. 
They either apply the divide-and-conquer strategy \cite{ji2017surfacenet} or is only applicable to synthetic data with low resolution inputs \cite{kar2017learning}. In contrast, our network focus on producing the depth map for one reference image at each time, which allows us to adaptively reconstruct a large scene directly.

\begin{figure}[t!]
  \centering
  \includegraphics[width=1\linewidth]{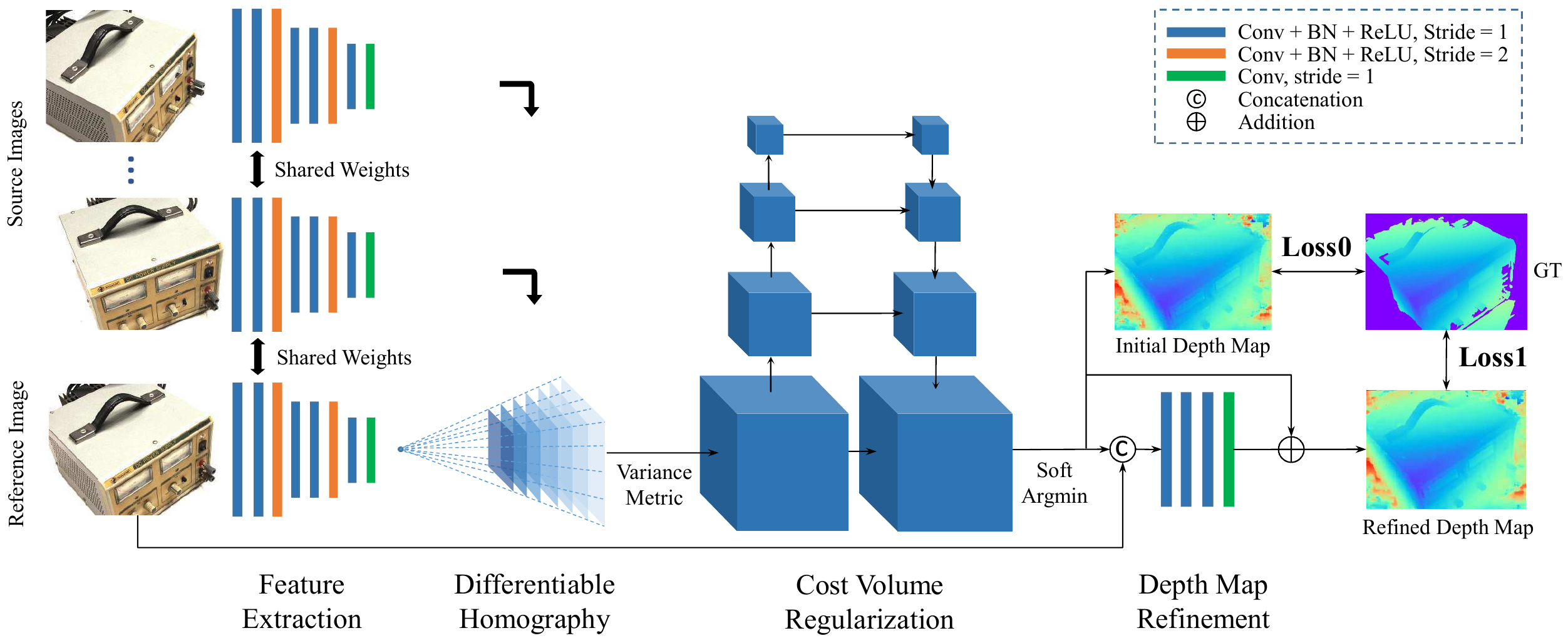}
  \caption{The network design of MVSNet. Input images will go through the 2D feature extraction network and the differentiable homograph warping to generate the cost volume. The final depth map output is regressed from the regularized probability volume and refined with the reference image}
  \label{fig:network}
\end{figure}
\vspace{-3mm}

\section{MVSNet}

This section describes the detailed architecture of the proposed network. The design of MVSNet strongly follows the rules of camera geometry and borrows the insights from previous MVS approaches. In following sections, we will compare each step of our network to the traditional MVS methods, and demonstrate the advantages of our learning-based MVS system. The full architecture of MVSNet is visualized in Fig. \ref{fig:network}.


\subsection{Image Features}

The first step of MVSNet is to extract the deep features $\{\mathbf{F}_i\}_{i=1}^{N}$ of the $N$ input images $\{\mathbf{I}_i\}_{i=1}^{N}$ for dense matching. An eight-layer 2D CNN is applied, where the strides of layer 3 and 6 are set to two to divide the feature towers into three scales. Within each scale, two convolutional layers are applied to extract the higher-level image representation. Each convolutional layer is followed by a batch-normalization (BN) layer and a rectified linear unit (ReLU) except for the last layer. Also, similar to common matching tasks, parameters are shared among all feature towers for efficient learning. 
	
The outputs of the 2D network are $N$ $32$-channel feature maps downsized by four in each dimension compared with input images. It is noteworthy that though the image frame is downsized after feature extraction, the original neighboring information of each remaining pixel has already been encoded into the 32-channel pixel descriptor, which prevents dense matching from losing useful context information. Compared with simply performing dense matching on original images, the extracted feature maps significantly boost the reconstruction quality (see Sec. \ref{sec:ablations}).

\subsection{Cost Volume}

The next step is to build a 3D cost volume from the extracted feature maps and input cameras. While previous works \cite{ji2017surfacenet,kar2017learning} divide the space using regular grids, for our task of depth map inference, we construct the cost volume upon the reference camera frustum. For simplicity, in the following we denote $\mathbf{I}_1$ as the reference image, $\{\mathbf{I}_i\}_{i=2}^{N}$ the source images, and $\{\mathbf{K}_i, \mathbf{R}_i, \mathbf{t}_i\}_{i=1}^{N}$ the camera intrinsics, rotations and translations that correspond to the feature maps.

\vspace{-3mm}
\subsubsection{Differentiable Homography} All feature maps are warped into different fronto-parallel planes of the reference camera to form $N$ feature volumes $\{\mathbf{V}_i\}_{i=1}^N$. The coordinate mapping from the warped feature map $\mathbf{V}_i(d)$ to $\mathbf{F}_i$ at depth $d$ is determined by the planar transformation $\mathbf{x'} \sim \mathbf{H}_i(d) \cdot \mathbf{x}$, where `$\sim$' denotes the projective equality and $\mathbf{H}_i(d)$ the homography between the $i^{th}$ feature map and the reference feature map at depth $d$. Let $\mathbf{n}_1$ be the principle axis of the reference camera, the homography is expressed by a $3 \times 3$ matrix:
	\begin{equation}
	\mathbf{H}_i(d) = \mathbf{K}_i \cdot \mathbf{R}_i \cdot \Big(\mathbf{I} - \frac{(\mathbf{t}_1 - \mathbf{t}_i) \cdot \mathbf{n}_1^T}{d} \Big) \cdot \mathbf{R}_1^T \cdot \mathbf{K}_1^T.
	\end{equation}
Without loss of generality, the homography for reference feature map $\mathbf{F}_1$ itself is an $3\times3$ identity matrix. The warping process is similar to that of the classical plane sweeping stereo \cite{collins1996space}, except that the differentiable bilinear interpolation is used to sample pixels from feature maps $\{\mathbf{F}_i\}_{i=1}^{N}$ rather than images $\{\mathbf{I}_i\}_{i=1}^{N}$. As the core step to bridge the 2D feature extraction and the 3D regularization networks, the warping operation is implemented in differentiable manner, which enables end-to-end training of depth map inference. 

\vspace{-3mm}
\subsubsection{Cost Metric} Next, we aggregate multiple feature volumes $\{\mathbf{V}_i\}_{i=1}^N$ to one cost volume $\mathbf{C}$. To adapt arbitrary number of input views, we propose a variance-based cost metric $\mathcal{M}$ for N-view similarity measurement.
	Let $W, H, D, F$ be the input image width, height, depth sample number and the channel number of the feature map, and $V = \frac{W}{4} \cdot \frac{H}{4} \cdot D \cdot F$ the feature volume size, our cost metric defines the mapping $\mathcal{M}: \underbrace{\mathbb{R}^V \times \cdots \times \mathbb{R}^V}_{N} \to \mathbb{R}^V$ that: 
	\begin{equation} \label{eq:metric}
	\mathbf{C} = \mathcal{M}(\mathbf{V}_1, \cdots, \mathbf{V}_N) = \frac{\sum\limits_{i=1}^N{(\mathbf{V}_i - \overbar{\mathbf{V}_i})^2}}{N}
	\end{equation}
Where $\overbar{\mathbf{V}_i}$ is the average volume among all feature volumes, and all operations above are element-wise. 
	
Most traditional MVS methods aggregate pairwise costs between the reference image and all source images in a heuristic way. Instead, our metric design follows the philosophy that all views should contribute equally to the matching cost and gives no preference to the reference image \cite{hartmann2017learned}. We notice that recent work \cite{hartmann2017learned} applies the mean operation with multiple CNN layers to infer the multi-patch similarity. Here we choose the `variance' operation instead because the `mean' operation itself provides no information about the feature differences, and their network requires pre- and post- CNN layers to help infer the similarity. In contrast, our variance-based cost metric explicitly measures the multi-view feature difference. In later experiments, we will show that such explicit difference measurement improves the validation accuracy.

\vspace{-3mm}
\subsubsection{Cost Volume Regularization} The raw cost volume computed from image features could be noise-contaminated (e.g., due to the existence of non-Lambertian surfaces or object occlusions) and should be incorporated with smoothness constraints to infer the depth map. Our regularization step is designed for refining the above cost volume $\mathbf{C}$ to generate a probability volume $\mathbf{P}$ for depth inference. Inspired by recent learning-based stereo \cite{kendall2017end} and MVS \cite{ji2017surfacenet,kar2017learning} methods, we apply the multi-scale 3D CNN for cost volume regularization. The four-scale network here is similar to a 3D version UNet \cite{ronneberger2015u}, which uses the encoder-decoder structure to aggregate neighboring information from a large receptive field with relatively low memory and computation cost. To further lessen the computational requirement, we reduce the 32-channel cost volume to 8-channel after the first 3D convolutional layer, and change the convolutions within each scale from 3 layers to 2 layers. The last convolutional layer outputs a 1-channel volume. We finally apply the \textit{softmax} operation along the depth direction for probability normalization. 
	
The resulting probability volume is highly desirable in depth map inference that it can not only be used for per-pixel depth estimation, but also for measuring the estimation confidence. We will show in Sec. \ref{sec:quality} that one can easily determine the depth reconstruction quality by analyzing its probability distribution, which leads to a very concise yet effective outlier filtering strategy in Sec. \ref{sec:filter}.

\subsection{Depth Map}

\subsubsection{Initial Estimation} 
\begin{figure}[t!]
  \centering
  \includegraphics[width=1\linewidth]{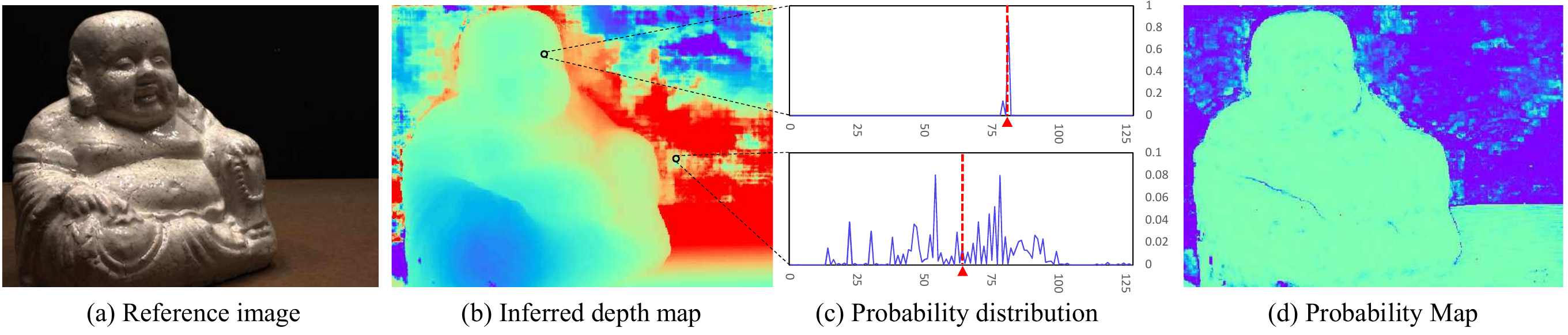}
  \caption{Illustrations on inferred depth map, probability distributions and probability map. (a) One reference image of \textit{scan} 114, \textit{DTU} dataset \cite{aanaes2016large}; (b) the inferred depth map; (c) the probability distributions of an inlier pixel (top) and an outlier pixel (bottom), where the x-axis is the index of depth hypothesis, y-axis the probability and red lines the soft argmin results; (d) the probability map. As shown in (c), the outlier's distribution is scattered and results in a low probability estimation in (d)}
  \label{fig:distribution}
\end{figure}

The simplest way to retrieve depth map $\mathbf{D}$ from the probability volume $\mathbf{P}$ is the pixel-wise winner-take-all \cite{collins1996space} (i.e., \textit{argmax}). 
However, the \textit{argmax} operation is unable to produce sub-pixel estimation, and cannot be trained with back-propagation due to its indifferentiability. Instead, we compute the \emph{expectation} value along the depth direction, i.e., the probability weighted sum over all hypotheses: 
	\begin{equation}
	\mathbf{D} = \sum\limits_{d=d_{min}}^{d_{max}} d \times \mathbf{P}(d)
	\end{equation}
Where $\mathbf{P}(d)$ is the probability estimation for all pixels at depth $d$. Note that this operation is also referred to as the \textit{soft argmin} operation in \cite{kendall2017end}. It is fully differentiable and able to approximate the argmax result. While the depth hypotheses are uniformly sampled within range $[d_{min}, d_{max}]$ during cost volume construction, the expectation value here is able to produce a continuous depth estimation. The output depth map (Fig. \ref{fig:distribution} (b)) is of the same size to 2D image feature maps, which is downsized by four in each dimension compared to input images.

\vspace{-3mm}
\subsubsection{Probability Map}\label{sec:quality} The probability distribution along the depth direction also reflects the depth estimation quality. Although the multi-scale 3D CNN has very strong ability to regularize the probability to the single modal distribution, we notice that for those falsely matched pixels, their probability distributions are scattered and cannot be concentrated to one peak (see Fig. \ref{fig:distribution} (c)). Based on this observation, we define the quality of a depth estimation $\hat{d}$ as the probability that the ground truth depth is within a small range near the estimation. As depth hypotheses are discretely sampled along the camera frustum, we simply take the probability sum over the four nearest depth hypotheses to measure the estimation quality. Notice that other statistical measurements, such as standard deviation or entropy can also be used here, but in our experiments we observe no significant improvement from these measurements for depth map filtering. Moreover, our probability sum formulation leads to a better control of thresholding parameter for outliers filtering. 

\vspace{-3mm}
\subsubsection{Depth Map Refinement} While the depth map retrieved from the probability volume is a qualified output, the reconstruction boundaries may suffer from oversmoothing due to the large receptive field involved in the regularization, which is similar to the problems in semantic segmentation \cite{chen2017deeplab} and image matting \cite{xu2017deep}. Notice that the reference image in natural contains boundary information, we thus use the reference image as a guidance to refine the depth map. Inspired by the recent image matting algorithm \cite{xu2017deep}, we apply a depth residual learning network at the end of MVSNet. The initial depth map and the resized reference image are concatenated as a 4-channel input, which is then passed through three 32-channel 2D convolutional layers followed by one 1-channel convolutional layer to learn the depth residual. The initial depth map is then added back to generate the refined depth map. The last layer does not contain the BN layer and the ReLU unit as to learn the negative residual. Also, to prevent being biased at a certain depth scale, we pre-scale the initial depth magnitude to range [0, 1], and convert it back after the refinement. 

\subsection{Loss}

Losses for both the initial depth map and the refined depth map are considered. We use the mean absolute difference between the ground truth depth map and the estimated depth map as our training loss. As ground truth depth maps are not always complete in the whole image (see Sec. \ref{sec:train}), we only consider those pixels with valid ground truth labels:
	\begin{equation}
	Loss = \sum\limits_{p \in \mathbf{p}_{valid}} \underbrace{\| d(p) - \hat{d_{i}}(p) \|_1}_{Loss0} + \lambda \cdot \underbrace{\| d(p) - \hat{d_{r}}(p) \|_1}_{Loss1}
	\end{equation}
Where $\mathbf{p}_{valid}$ denotes the set of valid ground truth pixels, $d(p)$ the ground truth depth value of pixel $p$, $\hat{d_{i}}(p)$ the initial depth estimation and $\hat{d_{r}}(p)$ the refined depth estimation. The parameter $\lambda$ is set to $1.0$ in experiments.

\section{Implementations}

\subsection{Training} \label{sec:train}

\subsubsection{Data Preparation} Current MVS datasets provide ground truth data in either point cloud or mesh formats, so we need to generate the ground truth depth maps ourselves. The \textit{DTU} dataset \cite{aanaes2016large} is a large-scale MVS dataset containing more than 100 scenes with different lighting conditions. As it provides the ground truth point cloud with normal information, we use the screened Poisson surface reconstruction (SPSR) \cite{kazhdan2013screened} to generate the mesh surface, and then render the mesh to each viewpoint to generate the depth maps for our training. The parameter, \textit{depth-of-tree} is set to 11 in SPSR to acquire the high quality mesh result. Also, we set the mesh \textit{trimming-factor} to 9.5 to alleviate mesh artifacts in surface edge areas. To fairly compare MVSNet with other learning based methods, we choose the same training, validation and evaluation sets as in SurfaceNet \cite{ji2017surfacenet}\footnote{Validation set: scans $\{$3, 5, 17, 21, 28, 35, 37, 38, 40, 43, 56, 59, 66, 67, 82, 86, 106, 117$\}$. Evaluation set: scans $\{$1, 4, 9, 10, 11, 12, 13, 15, 23, 24, 29, 32, 33, 34, 48, 49, 62, 75, 77, 110, 114, 118$\}$. Training set: the other 79 scans.}. Considering each scan contains 49 images with 7 different lighting conditions, by setting each image as the reference, \textit{DTU} dataset provides 27097 training samples in total.
	
\vspace{-3mm}
\subsubsection{View Selection} A reference image and two source images ($N=3$) are used in our training. We calculate a score $s(i, j) = \sum_{\mathbf{p}} \mathcal{G}(\theta_{ij}(\mathbf{p}))$ for each image pair according to the sparse points, where $\mathbf{p}$ is a common track in both view $i$ and $j$, $\theta_{ij}(\mathbf{p}) = (180/\pi)\arccos((\mathbf{c}_i - \mathbf{p}) \cdot (\mathbf{c}_j - \mathbf{p}))$ is $\mathbf{p}$'s baseline angle and $\mathbf{c}$ is the camera center. $\mathcal{G}$ is a piecewise Gaussian function \cite{zhang2015joint} that favors a certain baseline angle $\theta_0$:
\[ \mathcal{G}(\theta) =  \left\{
\begin{array}{ll}
      \exp(-\frac{(\theta - \theta_0)^2}{2\sigma_1^2}), \theta \leq \theta_0 \\
      \exp(-\frac{(\theta - \theta_0)^2}{2\sigma_2^2}), \theta > \theta_0 \\
\end{array} 
\right. \]
In the experiments, $\theta_0$, $\sigma_1$ and $\sigma_2$ are set to 5, 1 and 10 respectively. 

Notice that images will be downsized in feature extraction, plus the four-scale encoder-decoder structure in 3D regularization part, the input image size must be divisible by a factor of 32. Considering this requirement also the limited GPU memories, we downsize the image resolution from $1600\times1200$ to $800\times600$, and then crop the image patch with $W=640$ and $H=512$ from the center as the training input. The input camera parameters are changed accordingly. The depth hypotheses are uniformly sampled from $425mm$ to $935mm$ with a $2mm$ resolution ($D=256$). We use TensorFlow \cite{tensorflow2015-whitepaper} to implement MVSNet, and the network is trained on one Tesla P100 graphics card for around $100,000$ iterations. 

\begin{figure}[!t]
  \centering
  \includegraphics[width=1\linewidth]{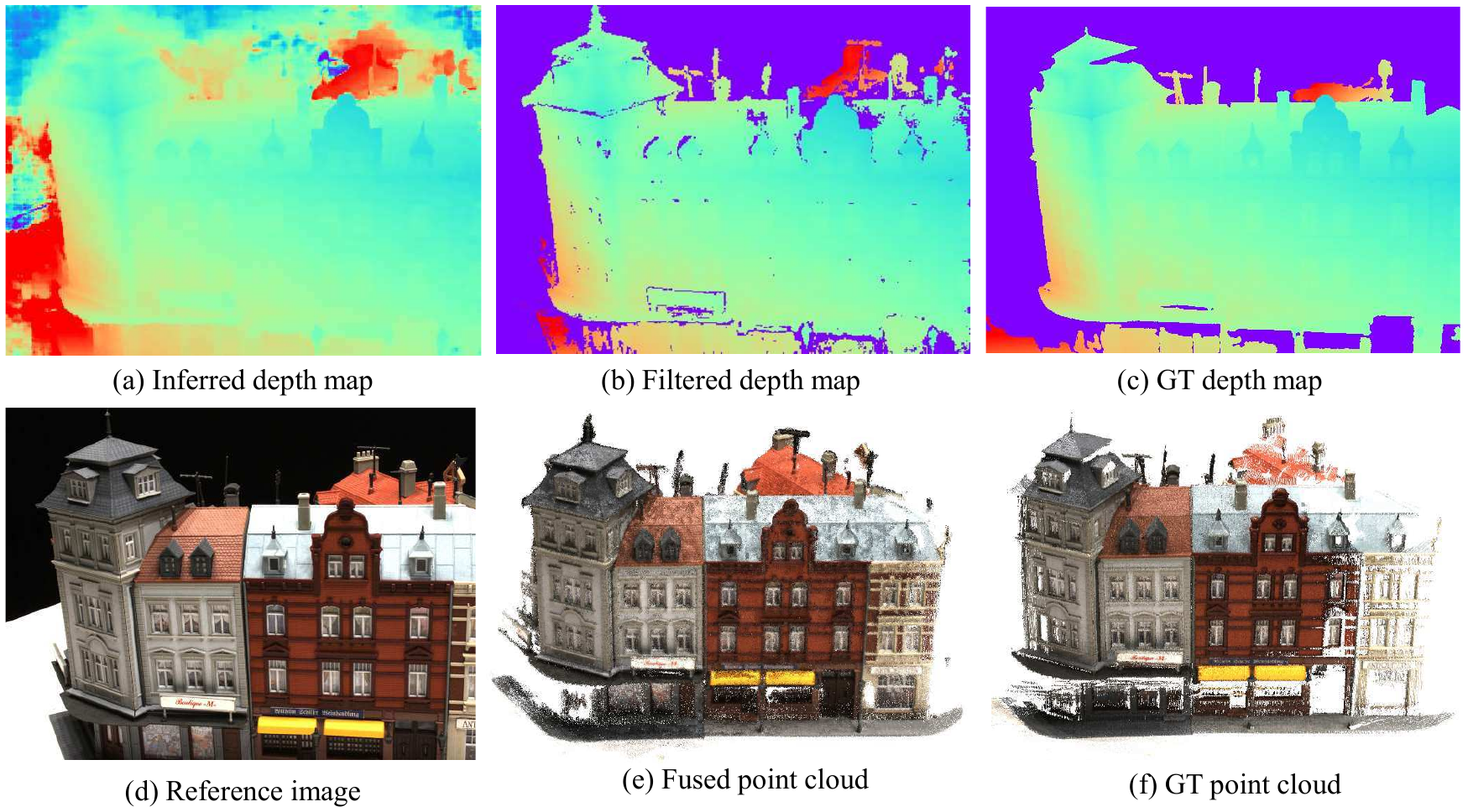}
  \caption{Reconstructions of \textit{scan} 9, \textit{DTU} dataset \cite{aanaes2016large}. From top left to bottom right: (a) the inferred depth map from MVSNet; (b) the filtered depth map after photometric and geometric filtering; (c) the depth map rendered from the ground truth mesh; (d) the reference image; (e) the final fused point cloud; (f) the ground truth point cloud}
  \label{fig:pipeline}
\end{figure}

\subsection{Post-processing}

\subsubsection{Depth Map Filter} \label{sec:filter} The above network estimates a depth value for every pixel. Before converting the result to dense point clouds, it is necessary to filter out outliers at those background and occluded areas. We propose two criteria, namely \textit{photometric} and \textit{geometric} consistencies for the robust depth map filtering. 
	
The photometric consistency measures the matching quality. As discussed in Sec. \ref{sec:quality}, we compute the probability map to measure the depth estimation quality. In our experiments, we regard pixels with probability lower than 0.8 as outliers. The geometric constraint measures the depth consistency among multiple views. Similar to the left-right disparity check for stereo, we project a reference pixel $p_1$ through its depth $d_1$ to pixel $p_i$ in another view, and then reproject $p_i$ back to the reference image by $p_i$'s depth estimation $d_i$. If the reprojected coordinate $p_{reproj}$ and and the reprojected depth $d_{reproj}$ satisfy $|p_{reproj} - p_1| < 1$ and $|d_{reproj} - d_1| / d_1 < 0.01$, we say the depth estimation $d_1$ of $p_1$ is two-view consistent. In our experiments, all depths should be at least three view consistent. This simple two-step filtering strategy shows strong robustness for filtering different kinds of outliers.

\subsubsection{Depth Map Fusion} Similar to other multi-view stereo methods \cite{galliani2015massively,schonberger2016pixelwise}, we apply a depth map fusion step to integrate depth maps from different views to a unified point cloud representation. The visibility-based fusion algorithm \cite{merrell2007real} is used in our reconstruction, where depth occlusions and violations across different viewpoints are minimized. To further suppress reconstruction noises, we determine the visible views for each pixel as in the filtering step, and take the average over all reprojected depths $\overbar{d_{reproj}}$ as the pixel's final depth estimation. The fused depth maps are then directly reprojected to space to generate the 3D point cloud. The illustration of our MVS reconstruction is shown in Fig. \ref{fig:pipeline}.

\section{Experiments}

\subsection{Benchmarking on \textit{DTU} dataset} \label{sec:dtu}

We first evaluate our method on the 22 evaluation scans of the \textit{DTU} dataset \cite{aanaes2016large}. The input view number, image width, height and depth sample number are set to $N=5$, $W=1600$, $H=1184$ and $D=256$ respectively. For quantitative evaluation, we calculate the \textit{accuracy} and the \textit{completeness} of both the distance metric \cite{aanaes2016large} and the percentage metric \cite{knapitsch2017tanks}. While the matlab code for the distance metric is given by \textit{DTU} dataset, we implement the percentage evaluation ourselves. Notice that the percentage metric also measures the overall performance of accuracy and completeness as the \textit{f-score}. To give a similar measurement for the distance metric, we define the \textit{overall score}, and take the average of mean accuracy and mean completeness as the reconstruction quality. 

\begin{figure}[!t]
  \centering
  \includegraphics[width=1\linewidth]{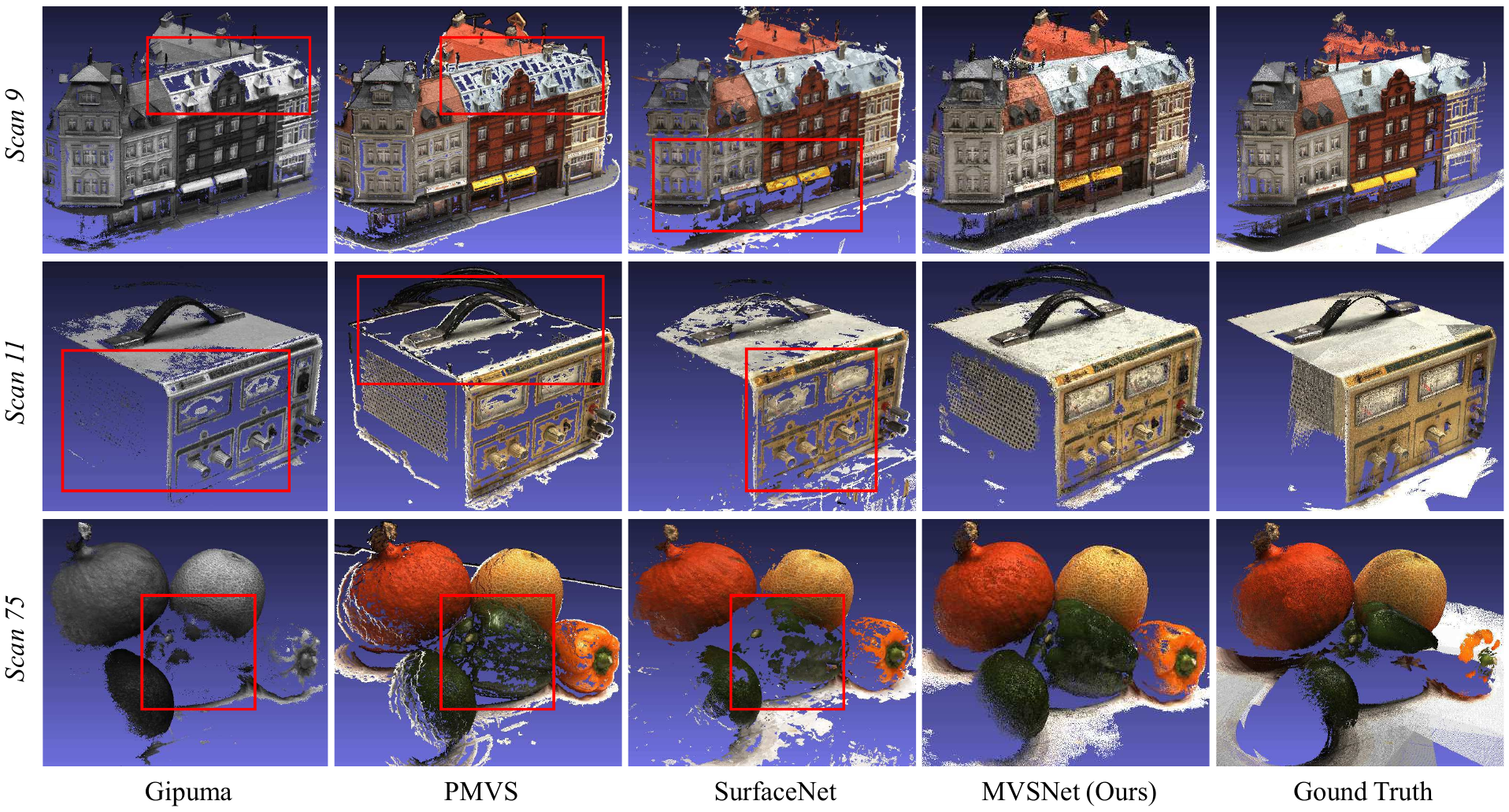}
  \caption{Qualitative results of \textit{scans} 9, 11 and 75 of \textit{DTU} dataset \cite{aanaes2016large}. Our MVSNet generates the most complete point clouds especially in those textureless and reflective areas. \textbf{Best viewed on screen}}
  \label{fig:dtu}
\end{figure}

Quantitative results are shown in Table \ref{table:dtu}. While Gipuma \cite{tola2012efficient} performs best in the accuracy, our MVSNet outperforms all methods in both the completeness and the overall quality \textbf{with a significant margin}. As shown in Fig. \ref{fig:dtu}, MVSNet produces the most complete point clouds especially in those textureless and reflected  areas, which are commonly considered as the most difficult parts to recover in MVS reconstruction.


\begin{table}[!t]
\centering
\caption{Quantitative results on the \textit{DTU}'s evaluation set \cite{aanaes2016large}. We evaluate all methods using both the distance metric \cite{aanaes2016large} (lower is better), and the percentage metric \cite{knapitsch2017tanks} (higher is better) with respectively $1mm$ and $2mm$ thresholds}
\resizebox{\textwidth}{!}{%
\begin{tabular}{c c c c | c c c | c c c}
\specialrule{.16em}{.08em}{.08em} 
           & \multicolumn{3}{c|}{Mean Distance (mm)}               & \multicolumn{3}{c|}{Percentage (\textless $1mm$)}  & \multicolumn{3}{c}{Percentage (\textless $2mm$)} \\ 
           & \multicolumn{3}{c|}{Acc. Comp.  \textit{overall}}     & \multicolumn{3}{c|}{Acc.           Comp.          \textit{f-score}}  & \multicolumn{3}{c}{Acc.           Comp.           \textit{f-score}} \\ \hline
Camp \cite{campbell2008using}       & 0.835          & 0.554          & 0.695          & 71.75          & 64.94          & 66.31          & 84.83          & 67.82          & 73.02         \\ 
Furu \cite{furukawa2010accurate}       & 0.613          & 0.941          & 0.777          & 69.55          & 61.52          & 63.26          & 78.99          & 67.88          & 70.93         \\ 
Tola \cite{tola2012efficient}      & 0.342          & 1.190          & 0.766          & 90.49          & 57.83          & 68.07          & 93.94          & 63.88          & 73.61         \\ 
Gipuma \cite{galliani2015massively}    & \textbf{0.283} & 0.873          & 0.578          & \textbf{94.65} & 59.93          & 70.64          & \textbf{96.42} & 63.81          & 74.16         \\ 
SurfaceNet\cite{ji2017surfacenet} & 0.450          & 1.04           & 0.745          & 83.8           & 63.38          & 69.95          & 87.15          & 67.99          & 74.4          \\ 
MVSNet (Ours)     & 0.396          & \textbf{0.527} & \textbf{0.462} & 86.46          & \textbf{71.13} & \textbf{75.69} & 91.06          & \textbf{75.31} & \textbf{80.25} \\ 
\specialrule{.16em}{.08em}{.081em}
\end{tabular}%
}
\label{table:dtu}
\end{table}

\begin{table}
\centering
\caption{Quantitative results on \textit{Tanks and Temples} benchmark \cite{knapitsch2017tanks}. MVSNet achieves best \textit{f-score} result among all submissions without any fine-tuning}
\label{tab:tt}
\resizebox{\textwidth}{!}{%
\begin{tabular}{lccccccccccc} \specialrule{.2em}{.1em}{.1em}
Method                      								& Rank  		& Mean    		& Family		& Francis 		& Horse 		& Lighthouse 	& M60   		& Panther 		& Playground 	& Train \\ \hline
MVSNet (Ours) 												& \textbf{3.00}	& \textbf{43.48}& 55.99			& 28.55			& 25.07			& 50.79			& \textbf{53.96}& \textbf{50.86}& 47.90			& 34.69 \\
Pix4D \cite{Pix4D}                       					& 3.12 			& 43.24 		& \textbf{64.45}& 31.91   		& \textbf{26.43}& 54.41      	& 50.58 		& 35.37   		& 47.78      	& 34.96 \\
COLMAP \cite{schonberger2016pixelwise}          			& 3.50      	& 42.14     	& 50.41  		& 22.25   		& 25.63 		& \textbf{56.43}& 44.83 		& 46.97   		& \textbf{48.53}& \textbf{42.04} \\
OpenMVG \cite{openMVG} + OpenMVS \cite{openMVS}				& 3.62      	& 41.71     	& 58.86  		& \textbf{32.59}& 26.25 		& 43.12      	& 44.73 		& 46.85   		& 45.97      	& 35.27 \\
OpenMVG \cite{openMVG} + MVE \cite{fuhrmann2014mve}         & 6.00      	& 38.00     	& 49.91  		& 28.19   		& 20.75 		& 43.35      	& 44.51 		& 44.76   		& 36.58      	& 35.95 \\
OpenMVG \cite{openMVG} + SMVS \cite{langguth2016shading}    & 10.38     	& 30.67     	& 31.93  		& 19.92   		& 15.02 		& 39.38      	& 36.51 		& 41.61   		& 35.89      	& 25.12 \\
OpenMVG-G \cite{openMVG} + OpenMVS \cite{openMVS}          	& 10.88     	& 22.86     	& 56.50  		& 29.63   		& 21.69 		& 6.55       	& 39.54 		& 28.48   		& 0.00       	& 0.53  \\
MVE \cite{fuhrmann2014mve}                         			& 11.25     	& 25.37     	& 48.59  		& 23.84   		& 12.70 		& 5.07       	& 39.62 		& 38.16   		& 5.81       	& 29.19 \\
OpenMVG \cite{openMVG} + PMVS \cite{furukawa2010accurate}   & 11.88     	& 29.66     	& 41.03  		& 17.70   		& 12.83 		& 36.68      	& 35.93 		& 33.20   		& 31.78      	& 28.10 \\
\specialrule{.2em}{.1em}{.1em}
\end{tabular}%
}
\end{table}
\begin{figure}[!t]
  \centering
  \includegraphics[width=1\linewidth]{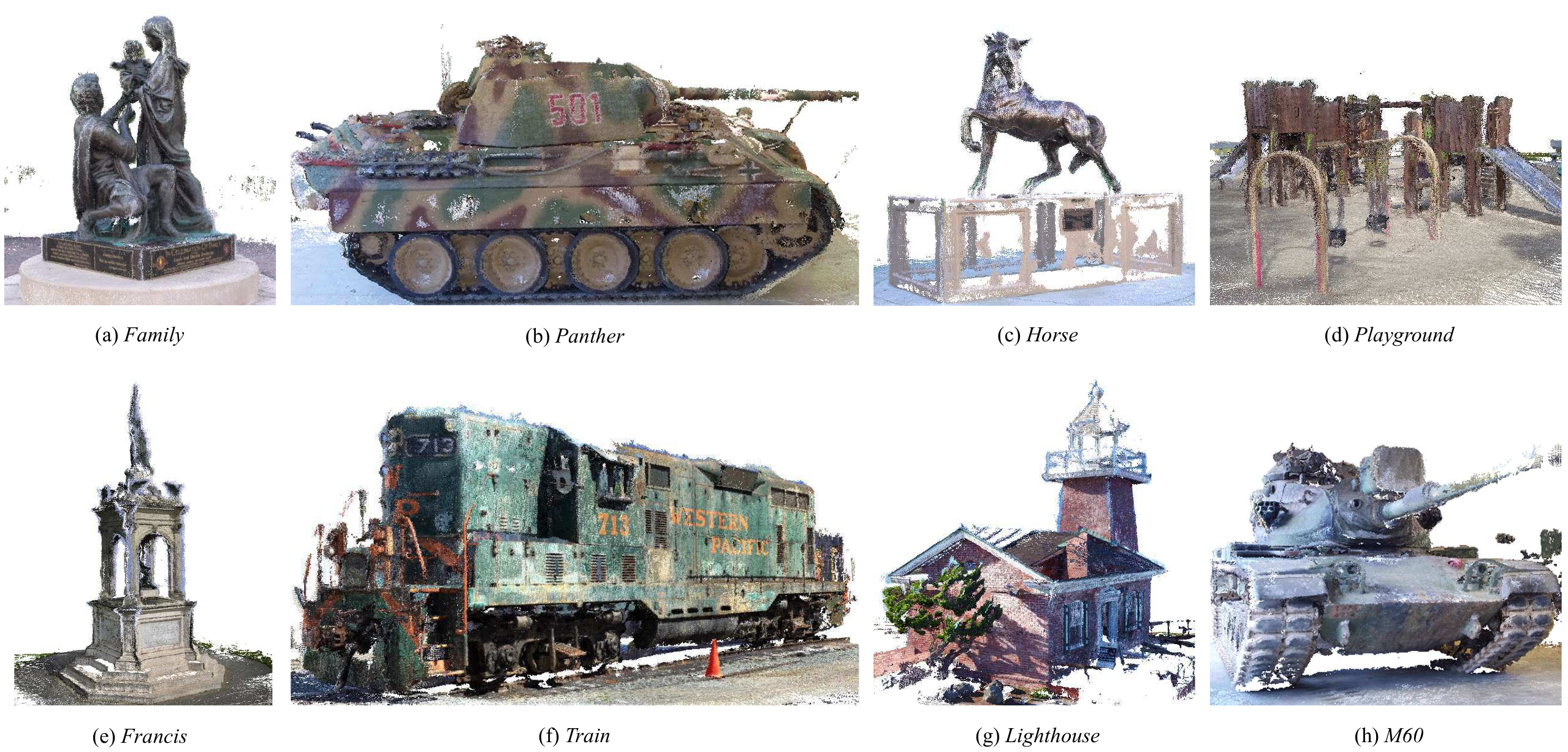}
  \caption{Point cloud results of the \textit{intermediate set} of \textit{Tanks and Temples} \cite{knapitsch2017tanks} dataset, which demonstrates the generalization power of MVSNet on complex outdoor scenes}
  \label{fig:intel}
\end{figure}

\subsection{Generalization on \textit{Tanks and Temples} dataset} \label{sec:tt}

The \textit{DTU} scans are taken under well-controlled indoor environment with fixed camera trajectory. To further demonstrate the generalization ability of MVSNet, we test the proposed method on the more complex outdoor \textit{Tanks and Temples} dataset \cite{knapitsch2017tanks}, using the model trained on \textit{DTU} \textbf{without any fine-tuning}. While we choose $N=5$, $W=1920$, $H=1056$ and $D=256$ for all reconstructions, the depth range and the source image set for the reference image are determined according to sparse point cloud and camera positions, which are recovered by the open source SfM software OpenMVG \cite{openMVG}.

Our method ranks first before April 18, 2018 among all submissions of the \textit{intermediate set} \cite{knapitsch2017tanks} according to the online benchmark (Table \ref{tab:tt}). Although the model is trained on the very different \textit{DTU} indoor dataset, MVSNet is still able to produce the best reconstructions on these outdoor scenes, demonstrating the strong generalization ability of the proposed network. The qualitative point cloud results of the \textit{intermediate set} are visualized in Fig. \ref{fig:intel}.

\subsection{Ablations} \label{sec:ablations}

This section analyzes several components in MVSNet. For all following studies, we use the validation loss to measure the reconstruction quality. The 18 validation scans (see Sec. \ref{sec:train}) are pre-processed as the training set that we set $N=3$, $W=640$, $H=512$ and $D=256$ for the validation loss computation. 

\begin{figure}[!t]
  \centering
  \includegraphics[width=1\linewidth]{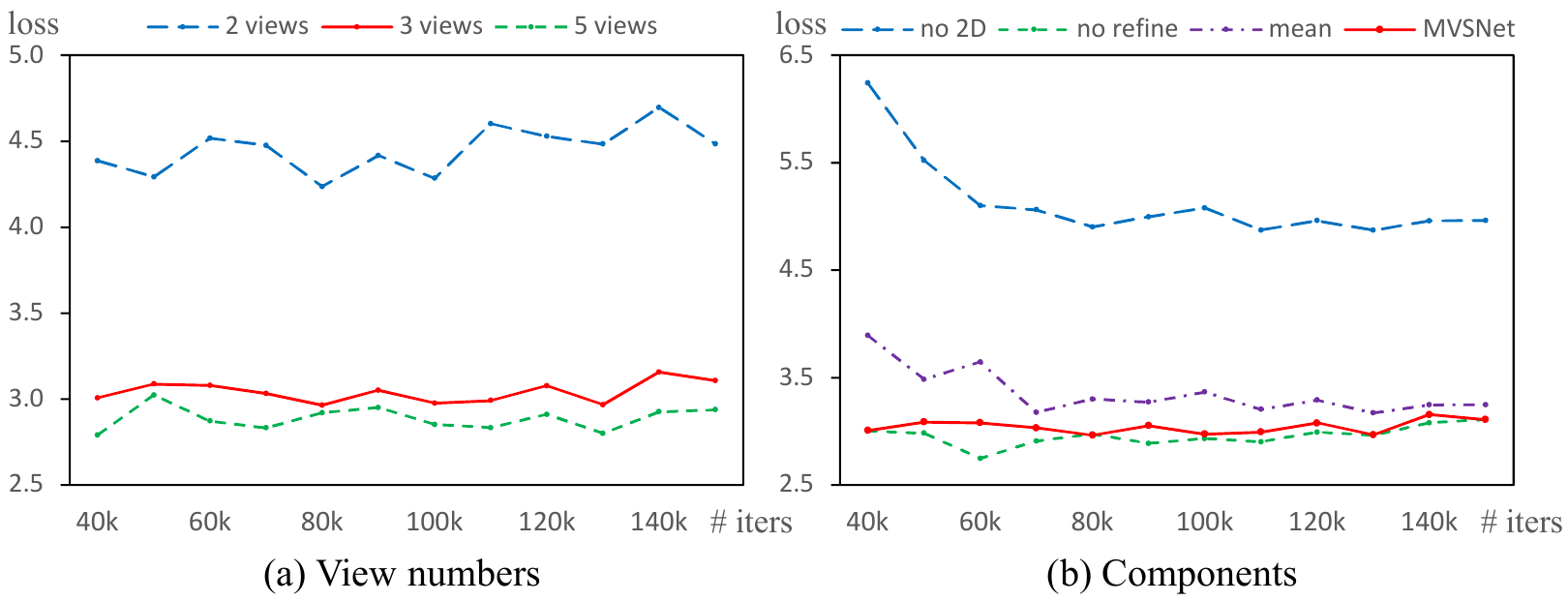}
  \caption{Ablation studies. (a) Validation losses of different input view numbers. (b) Ablations on 2D image feature, cost metric and depth map refinement
  }
  \label{fig:ablations}
\end{figure}

\subsubsection{View Number}
We first study the influence of the input view number $N$ and demonstrate that our model can be applied to arbitrary views of input. While the model in Sec. \ref{sec:train} is trained using $N = 3$ views, we test the model using $N = 2, 3, 5$ respectively. As expected, it is shown in Fig. \ref{fig:ablations} (a) that adding input views can lower the validation loss, which is consistent with our knowledge about MVS reconstructions. It is noteworthy that testing with $N=5$ performs better than with $N=3$, even though the model is trained with the 3 views setting. This highly desirable property makes MVSNet flexible enough to be applied the different input settings.

\vspace{-3mm}
\subsubsection{Image Features}
We demonstrate in this study that the learning based image feature could significantly boost the MVS reconstruction quality. To model the traditional patch-based image feature in MVSNet, we replace the original 2D feature extraction network with a single 32-channel convolutional layer. The filter kernel is set to a large number of $7\times7$ and the stride is set to 4. As shown in Fig. \ref{fig:ablations} (b), network with the 2D feature extraction significantly outperforms the single layer one on validation loss.

\vspace{-3mm}
\subsubsection{Cost Metric}
We also compare our variance operation based cost metric with the mean operation based metric \cite{hartmann2017learned}. The element-wise variance operation in Eq. \ref{eq:metric} is replaced with the mean operation to train the new model. It can be found in Fig. \ref{fig:ablations} (b) that our cost metric results in a faster convergence with lower validation loss, which demonstrates that it is more reasonable to use the explicit difference measurement to compute the multi-view feature similarity.

\vspace{-3mm}
\subsubsection{Depth Refinement} 
Lastly, we train MVSNet with and without the depth map refinement network. The models are also tested on \textit{DTU} evaluation set as in Sec. \ref{sec:dtu}, and we use the percentage metric \cite{knapitsch2017tanks} to quantitatively compare the two models. While Fig. \ref{fig:ablations} (b) shows that the refinement does not affect the validation loss too much, the refinement network improves the evaluation results from 75.58 to 75.69 ($<1mm$ \textit{f-score}) and from 79.98 to 80.25 ($<2mm$ \textit{f-score}). 

\subsection{Discussions}

\subsubsection{Running Time}
We compare the running speed of MVSNet to Gipuma \cite{galliani2015massively}, COLMAP \cite{schonberger2016pixelwise} and SurfaceNet \cite{ji2017surfacenet} using the $DTU$ evaluation set. The other methods are compiled from their source codes and all methods are tested in the same machine. MVSNet is much more efficient that it takes around 230 seconds to reconstruct one scan (\textbf{4.7 seconds} per view). The running speed is $\sim5\times$ faster than Gipuma, $\sim100\times$ than COLMAP and $\sim160\times$ than SurfaceNet. 

\vspace{-3mm}
\subsubsection{GPU Memory}
The GPU memory required by MVSNet is related to the input image size and the depth sample number. In order to test on the \textit{Tanks and Temples} with the original image resolution and sufficient depth hypotheses, we choose the Tesla P100 graphics card (16 GB) to implement our method. It is noteworthy that the training and validation on \textit{DTU} dataset could be done using one consumer level GTX 1080ti graphics card (11 GB).

\vspace{-3mm}
\subsubsection{Training Data}
As mentioned in Sec. \ref{sec:train}, \textit{DTU} provides ground truth point clouds with normal information so that we can convert them into mesh surfaces for depth maps rendering. However, currently \textit{Tanks and Temples} dataset does not provide the normal information or mesh surfaces, so we are unable to fine-tune MVSNet on \textit{Tanks and Temples} for better performance. 

Although using such rendered depth maps have already achieved satisfactory results, some limitations still exist: 1) the provided ground truth meshes are not $100\%$ complete, so some triangles behind the foreground will be falsely rendered to the depth map as the valid pixels, which may deteriorate the training process. 2) If a pixel is occluded in all other views, it should not be used for training. However, without the complete mesh surfaces we cannot correctly identify the occluded pixels. We hope future MVS datasets could provide ground truth depth maps with complete occlusion and background information.

\section{Conclusion}
We have presented a deep learning architecture for MVS reconstruction. The proposed MVSNet takes unstructured images as input, and infers the depth map for the reference image in an end-to-end fashion. The core contribution of MVSNet is to encode the camera parameters as the differentiable homography to build the cost volume upon the camera frustum, which bridges the 2D feature extraction and 3D cost regularization networks. It has been demonstrated on \textit{DTU} dataset that MVSNet not only significantly outperforms previous methods, but also is more efficient in speed by several times. Also, MVSNet have produced the state-of-the-art results on \textit{Tanks and Temples} dataset without any fine-tuning, which demonstrates its strong generalization ability.

\bibliographystyle{splncs04}
\bibliography{egbib}

\end{document}